\documentclass[a4paper,fleqn]{cas-dc}
\usepackage{newtxtext} 
\usepackage{algorithm}
\usepackage{algpseudocode}

\usepackage[numbers,square,sort&compress]{natbib}
\usepackage{hyperref}
\hypersetup{
    colorlinks=true,
    linkcolor=blue,      
    filecolor=blue,      
    urlcolor=blue,       
    citecolor=blue,      
}

\def\tsc#1{\csdef{#1}{\textsc{\lowercase{#1}}\xspace}}
\tsc{WGM}
\tsc{QE}

\begin{document}
\let\WriteBookmarks\relax
\def\floatpagepagefraction{1}
\def\textpagefraction{.001}

\shorttitle{}    

\shortauthors{}  

\title [mode = title]{Decentralized LoRA Augmented Transformer with Context-aware Multi-scale Feature Learning for Secured Eye Diagnosis}  

\author[label1, label2]{Md. Naimur Asif Borno}
\affiliation[label1]{organization={Research Assistant, The University of Queensland},
            addressline={308 Queen St}, 
            city={Brisbane City},
            postcode={QLD 4000}, 
            state={Queensland},
            country={Australia}}
\affiliation[label2]{organization={Mechatronics Engineering, Rajshahi University of Engineering \& Technology},
            city={Rajshahi},
            postcode={6204}, 
            country={Bangladesh}}

\author[label3,label4]{Md Sakib Hossain Shovon}
\affiliation[label3]{organization={Researcher, The University of Queensland},
            addressline={308 Queen St}, 
            city={Brisbane City},
            postcode={QLD 4000}, 
            state={Queensland},
            country={Australia}}
\affiliation[label4]{organization={Department of Computer Science, American International University Bangladesh},
            city={Dhaka},
            postcode={1216}, 
            country={Bangladesh}}
            
\author[label5]{MD Hanif Sikder} 
\affiliation[label5]{organization={Department of Computer Science, University of South Asia-Bangladesh},
            city={Dhaka},
            postcode={1216},
            country={Bangladesh}}          
\author[label6]{Iffat Firozy Rimi}
\affiliation[label6]{organization={Department of Computer Science and Engineering, Daffodil International University},
            city={Dhaka},
            country={Bangladesh}}
\author[label7]{Tahani Jaser Alahmadi}
\affiliation[label7]{Department of Information Systems, College of Computer and Information Sciences, Princess Nourah bint Abdulrahman University, P.O. Box 84428, Riyadh, Saudi Arabia. Email: tjalahmadi@pnu.edu.sa}

\author[label8, label9]{Mohammad Ali Moni \corref{cor1}} 
\affiliation [label8]{organization={Faculty of Health, Medicine and Behavioural Sciences, The University of Queensland},
            addressline={308 Queen St}, 
            city={Brisbane City},
            postcode={QLD 4000}, 
            state={Queensland},
            country={Australia}}
\affiliation [label9]{organization={AI \& Digital Health Technology Artificial Intelligence and Cyber Futures Institute Charles Sturt University},
            addressline={308 Queen St}, 
            city={Bathurst NSW},
            country={Australia}}

\tnotetext[1]{Authors 1808009@student.ruet.ac.bd (Md.N.A. Borno),and sakib.aiub.cs@gmail.com (M.S.H. Shovon) contributed equally.}
\cortext[cor1]{Corresponding author email address: m.moni@uq.edu.au (M.A. Moni)}
\begin{abstract}
Accurate and privacy-preserving diagnosis of ophthalmic diseases remains a critical challenge in medical imaging, particularly given the limitations of existing deep learning models in handling data imbalance, data privacy concerns, spatial feature diversity, and clinical interpretability. This paper proposes a novel Data-efficient Image Transformer (DeiT)-based framework that integrates context-aware multiscale patch embedding, Low-Rank Adaptation (LoRA), knowledge distillation, and federated learning to address these challenges in a unified manner. The proposed model effectively captures both local and global retinal features by leveraging multi-scale patch representations with local and global attention mechanisms. LoRA integration enhances computational efficiency by reducing the number of trainable parameters, while federated learning ensures secure, decentralized training without compromising data privacy. A knowledge distillation strategy further improves generalization in data-scarce settings. Comprehensive evaluations on two benchmark datasets—OCTDL and the Eye Disease Image Dataset—demonstrate that the proposed framework consistently outperforms both traditional CNNs and state-of-the-art transformer architectures across key metrics including AUC, F1 score, and precision. Furthermore, Grad-CAM++ visualizations provide interpretable insights into model predictions, supporting clinical trust. This work establishes a strong foundation for scalable, secure, and explainable AI applications in ophthalmic diagnostics.
\end{abstract}




\maketitle

\section{Introduction}
Vision impairment remains a critical public health issue, profoundly affecting an individual’s autonomy, educational attainment, socioeconomic participation, and overall quality of life. Beyond its clinical burden, vision loss imposes a substantial economic cost on healthcare systems and societies. The human eye—housing more than half of the body’s sensory receptors—is particularly susceptible to degenerative and chronic conditions such as age-related macular degeneration (AMD), diabetic macular edema (DME), glaucoma, and diabetic retinopathy (DR). These diseases often progress silently and, if not detected and treated early, can result in irreversible blindness \cite{bakkouri2025ugs, rehman2018anatomy, kandel2022quality, saranya2025genetic}. Timely, accurate, and scalable diagnostic systems are therefore essential for effective intervention and long-term visual preservation.

In recent years, deep learning (DL) has markedly transformed medical image analysis, delivering substantial improvements in tasks such as segmentation, classification, and anomaly detection. These advancements have proven invaluable in supporting clinicians with the early and precise diagnosis of ocular diseases. Among DL techniques, Convolutional Neural Networks (CNNs) have been widely adopted in ophthalmology, particularly for the automated detection of retinal disorders, owing to their robust capability in extracting local image features \cite{csener2023classification}. For instance, \cite{shamsan2023automatic} proposed a hybrid approach that integrates feature extraction and fusion strategies to classify eye disease images. Similarly, \cite{muthukannan2022optimized} employed the Flower Pollination Optimization Algorithm (FPOA) in conjunction with a CNN model for ophthalmic disease classification, achieving a classification accuracy of 95.27\%. Despite their success, CNNs are inherently constrained by their architectural inductive biases and localized receptive fields. These limitations impair their ability to model long-range dependencies and global spatial relationships, which are essential for comprehensively understanding intricate retinal abnormalities \cite{martin2025evaluation, iratni2025transformers, sikder2024weighted}.

To address these limitations, transformer-based architectures—originally developed for natural language processing—have been adapted for vision tasks. Vision Transformers (ViTs) introduce self-attention mechanisms that enable global context modeling, offering a compelling alternative to CNNs for medical imaging tasks. In ophthalmology, ViTs have demonstrated notable success in fundus and optical coherence tomography (OCT) image analysis, showing superior performance over conventional CNNs in several classification and segmentation tasks \cite{jiang2022computer}. For example, Pyramid Vision Transformers (PVT) outperformed baseline ViT and CNN models in detecting glaucoma, DR, and cataracts, achieving classification accuracies above 85\% \cite{radhakrishnan2025eye}. Furthermore, \cite{he2023interpretable} improved a transformer-based model to achieve an impressive F1 score of 97.10\%.  These results underscore the promise of transformer-based approaches in capturing subtle, spatially distributed disease markers.

Despite these advancements, standalone transformer models remain constrained by high computational demands, limited generalizability in low-data settings, and suboptimal performance in preserving local texture information. Convolutional Vision Transformers (CvTs), while introducing early-stage convolutions to compensate for these shortcomings, achieved only 68\% accuracy on the ODIR-5k dataset and exhibited poor handling of class imbalance and fine-grained features \cite{tuwan2025eye}. Similarly, hybrid CNN-transformer models such as EfficientNet have shown competitive performance but often obscure clinically relevant details, particularly in high-resolution retinal images. Likewise, \cite{agarwal2025hdl} employed a CNN augmented with ant colony optimization to improve computational efficiency, yet the addition of a transformer-based feature extraction module introduced considerable computational overhead. 

Emerging transformer variants like Swin Transformers and OWL-ViTs have introduced hierarchical attention mechanisms to improve local-global feature integration, with some achieving accuracies as high as 99.4\% for glaucoma detection \cite{sivakumar2025enhancing}. However, these architectures tend to be computationally intensive, making them impractical for deployment in resource-constrained clinical environments. Furthermore, many of these approaches assume access to large-scale, well-labeled datasets—a significant barrier in medical imaging, where expert annotation is both time-consuming and expensive.

Data-Efficient Image Transformers (DeiT) address these critical gaps by offering a lightweight, computationally accessible alternative to traditional ViTs, while maintaining competitive performance through knowledge distillation \cite{touvron2021training}. In this framework, a compact transformer (student) is trained to replicate the outputs of a powerful CNN-based model (teacher), thereby transferring domain-relevant knowledge while significantly reducing data and compute requirements. This makes DeiT particularly attractive for ophthalmic diagnostics.

Despite the advantages of Data-efficient Image Transformers (DeiT), several domain-specific challenges remain unresolved in ophthalmic applications. First, retinal diseases often present through subtle and spatially localized indicators—such as microaneurysms, hard exudates, and changes in the optic disc—which standard patch embedding mechanisms may fail to capture effectively. Second, class imbalance is a persistent problem in retinal imaging datasets. Rare yet clinically critical conditions are frequently underrepresented, leading to biased model predictions and diminished diagnostic performance for minority classes. For instance, although \cite{ouda2022multiple} incorporated extensive preprocessing techniques, their approach did not adequately address the underlying issue of data imbalance. Similarly, \cite{sharma2024efficient} applied image filtering using a Gaussian blur algorithm, but also failed to mitigate this critical limitation. Third, the centralized architecture of most current AI pipelines raises significant concerns regarding data privacy and governance—especially in medical domains governed by strict regulatory frameworks. Patient data is typically siloed across institutions, complicating inter-organizational data sharing and collaborative model development. While encryption-based privacy-preserving solutions have been proposed \cite{kamal2021new}, their high computational demands and limited integration with contemporary deep learning frameworks significantly hinder their real-world applicability.

To mitigate these concerns, federated learning (FL) has emerged as a privacy-preserving solution, enabling collaborative model training across distributed clients without sharing raw data \cite{qu2022rethinking}. FL facilitates compliance with data governance regulations while maintaining model generalizability across diverse patient populations. Integrating transformer architectures into FL frameworks presents a promising research direction but remains underexplored—particularly with regard to DeiT, which is well-suited for decentralized training due to its low computational footprint. A recent study by \cite{kaissis2021end} introduced encrypted inference into FL for improved data confidentiality, but lacked integration with data-efficient transformer models and did not address challenges related to class imbalance or model interpretability.

Furthermore, most existing frameworks fall short in providing clinically interpretable outputs, a prerequisite for real-world adoption. Techniques like Bayesian modeling (e.g., BayesEG) have improved uncertainty quantification but lack spatial localization capabilities critical for clinical decision support \cite{gao2023bayeseg}. Gradient-weighted Class Activation Mapping (Grad-CAM) was utilized in \cite{jaimes2025detection} to visualize the model’s decision-making process; however, it falls short in providing interpretability of the intermediate layers’ representations. In contrast, Grad-CAM++ offers high-resolution visual explanations by identifying image regions most influential to a model’s decision, making it a powerful tool for validating AI predictions against known pathological features \cite{soomro2024grad++}. However, few transformer-based models—particularly in federated or privacy-sensitive contexts—leverage such interpretability tools effectively.

In sum, while prior research has individually addressed several key issues—improving feature extraction through transformers \cite{pu2024advantages}, addressing class imbalance via sampling techniques \cite{zhuang2023class}, and preserving data privacy using federated learning \cite{kaushal2023eye}—there remains no unified, domain-optimized framework that holistically integrates these components. The current landscape lacks a scalable, interpretable, and privacy-aware solution that is also computationally efficient and suitable for deployment across diverse clinical settings.

To tackle the key challenges in ophthalmic disease diagnosis—namely limited labeled data, privacy concerns, model interpretability, and class imbalance—we propose a comprehensive and data-efficient framework based on the DeiT architecture. At the core of our model is a context-aware multiscale patch embedding strategy that extracts two types of image patches (16×16 and 32×32) to capture both fine local structures and broader global context. To further enhance spatial representation, we apply local window attention to the smaller patches and global self-attention to the larger patches before feeding them into parallel transformer encoders. This allows the model to learn both detailed and holistic visual patterns that are critical for accurately identifying retinal abnormalities. To address the issue of class imbalance commonly found in medical datasets, we introduce stratified and weighted random sampling, which ensures that rare but clinically important conditions are better represented during training. Our framework also incorporates Low-Rank Adaptation (LoRA) to reduce the number of trainable parameters while maintaining performance, making the model more suitable for environments with limited data. In addition, we adopt a knowledge distillation setup where a high-capacity teacher model guides a smaller student model, improving learning efficiency and generalization. To ensure patient data privacy, we implement a FL approach that allows training across decentralized institutions without sharing sensitive data. Finally, we integrate Grad-CAM++ to generate visual explanations of the model’s predictions, highlighting disease-relevant regions and improving transparency for clinical use. Collectively, these contributions form a well-rounded, scalable, and interpretable framework for privacy-preserving AI in ophthalmic disease classification. 
\begin{figure*}[!h]
    \centering
    \includegraphics[width=1\linewidth]{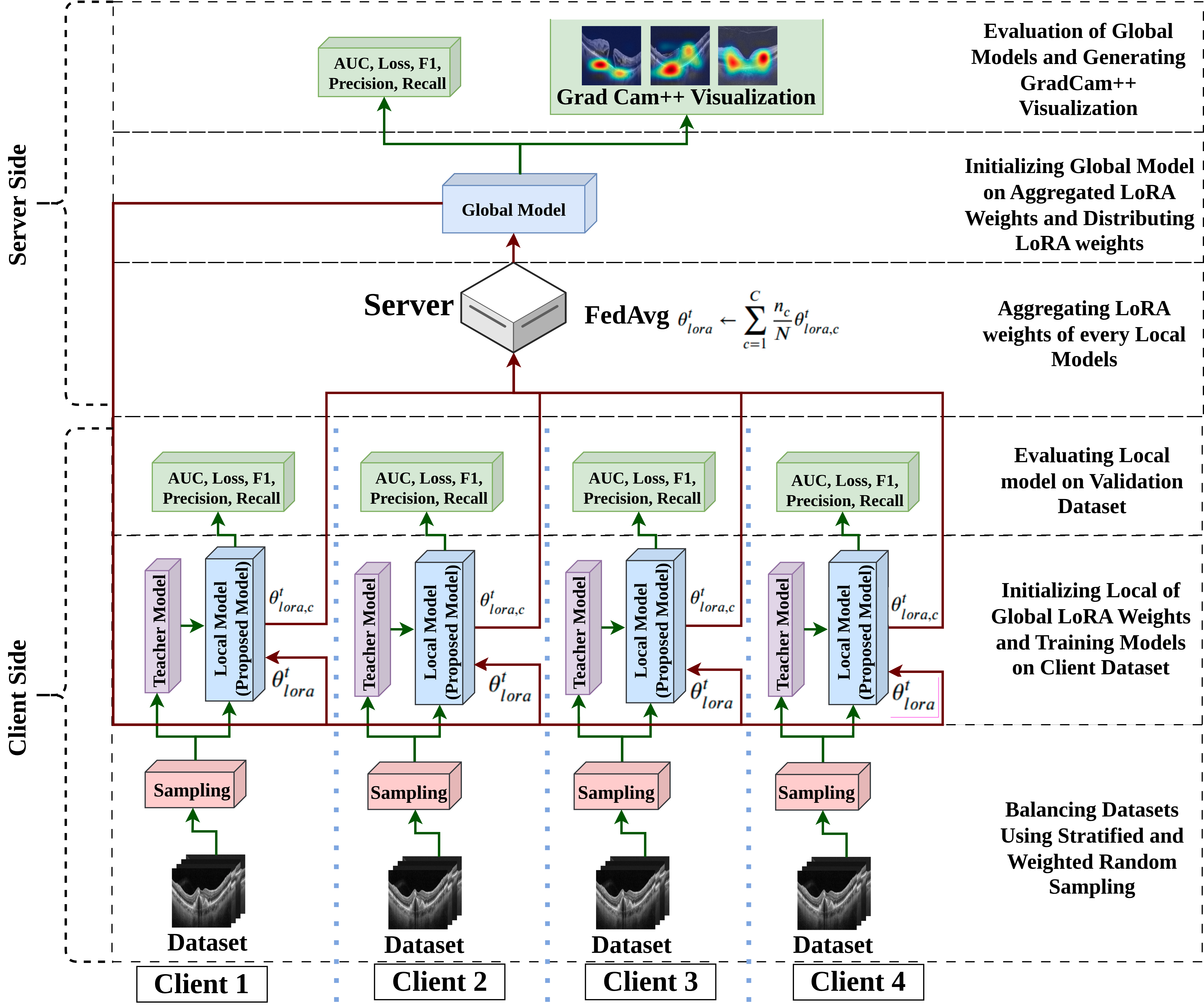}
    \caption{Framework of the proposed Local Model in a decentralized environment. Each Local Model is initialized with the LoRA weights from the Global Model and trained independently on client data. The Global Model aggregates the LoRA updates from all Local Models for collaborative learning. A detailed architecture of the Local Model is shown in \hyperref[fig:Figure_2]{Figure \ref*{fig:Figure_2}(b)}.}
    \label{fig:Figure_1}
\end{figure*}
\begin{table*}[!ht]
\centering
\caption{\bf Overview of the OCTDl and the Eye Disease Dataset(EDD).}
\begin{tabular}{p{1cm}lcccc}
 \hline
\textbf{Dataset} & \textbf{Class} & \textbf{Train} & \textbf{Test} & \textbf{Validation} & \textbf{Class ID}\\
\hline
\multirow{8}{*}{\rotatebox{90}{OCTDL Dataset}} 
& Age-related Macular Degeneration(AMD) & 985  & 123  & 123 & 0 \\
& Diabetic Macular Edema(DME)          & 119  & 14   & 14  & 1 \\
& Epiretinal Membrane(ERM)             & 125  & 15   & 15  & 2 \\
& Normal(NO)                           & 266  & 33   & 33  & 3 \\
& Retinal Artery Occlusion(RAO)        & 14   & 4    & 4   & 4 \\
& Retinal Vein Occlusion(RVO)          & 81   & 10   & 10  & 5 \\
& Vitreomacular Interface Disease(VID) & 62   & 7    & 7   & 6 \\
& \textbf{Total(OCTDL)}                & \textbf{1,652} & \textbf{206} & \textbf{206} \\
\hline
\multirow{8}{*}{\rotatebox{90}{Eye Disease Dataset}} 
& Central Serous Chorioretinopathy      & 81   & 10   & 10  & 0  \\
& Disc Edema                            & 103  & 12   & 12  & 1 \\
& Macular Scar                          & 356  & 44   & 44  & 2 \\
& Myopia                                & 407  & 50   & 50  & 3 \\
& Pterygium                             & 15   & 1    & 1   & 4 \\
& Retinal Detachment                    & 101  & 12   & 12  & 5 \\
& Retinitis Pigmentosa                  & 113  & 13   & 13  & 6 \\
& \textbf{Total (Eye Disease)}          & \textbf{1,176} & \textbf{142} & \textbf{142} \\
\hline
\end{tabular}
\label{table:Table_1}
\end{table*}
\section{Methodology}
\subsection{Dataset Descriptions}
Two main datasets were used for training and evaluating our model: the Optical Coherence Tomography Dataset for Image-Based Deep Learning Methods (OCTDL)\citep{Kulyabin2024-hk}  and the Eye Disease-Image Dataset(EDD)\citep{Khatun2024-je}. The OCTDL dataset contains over 2,000 high-resolution OCT images, each labeled according to specific disease categories and retinal pathologies. For this study, the dataset was split into training (1,652 images), testing (206 images), and validation (206 images) subsets. Various techniques such as resizing, random horizontal flips, random rotations, and normalization were applied to improve the model's robustness and generalization. The Eye Disease Image Dataset includes 1,460 images, categorized by condition, and was also divided into training (1,176 images), testing (142 images), and validation (142 images) sets. Consistent augmentation strategies were used to enhance the model's performance and adaptability. \hyperref[table:Table_1]{Table \ref*{table:Table_1}} presents an overview of the two datasets, including the number of images and the corresponding Class IDs used in this study.
\begin{figure*}[!ht]
    \centering
    \includegraphics[width=1\linewidth]{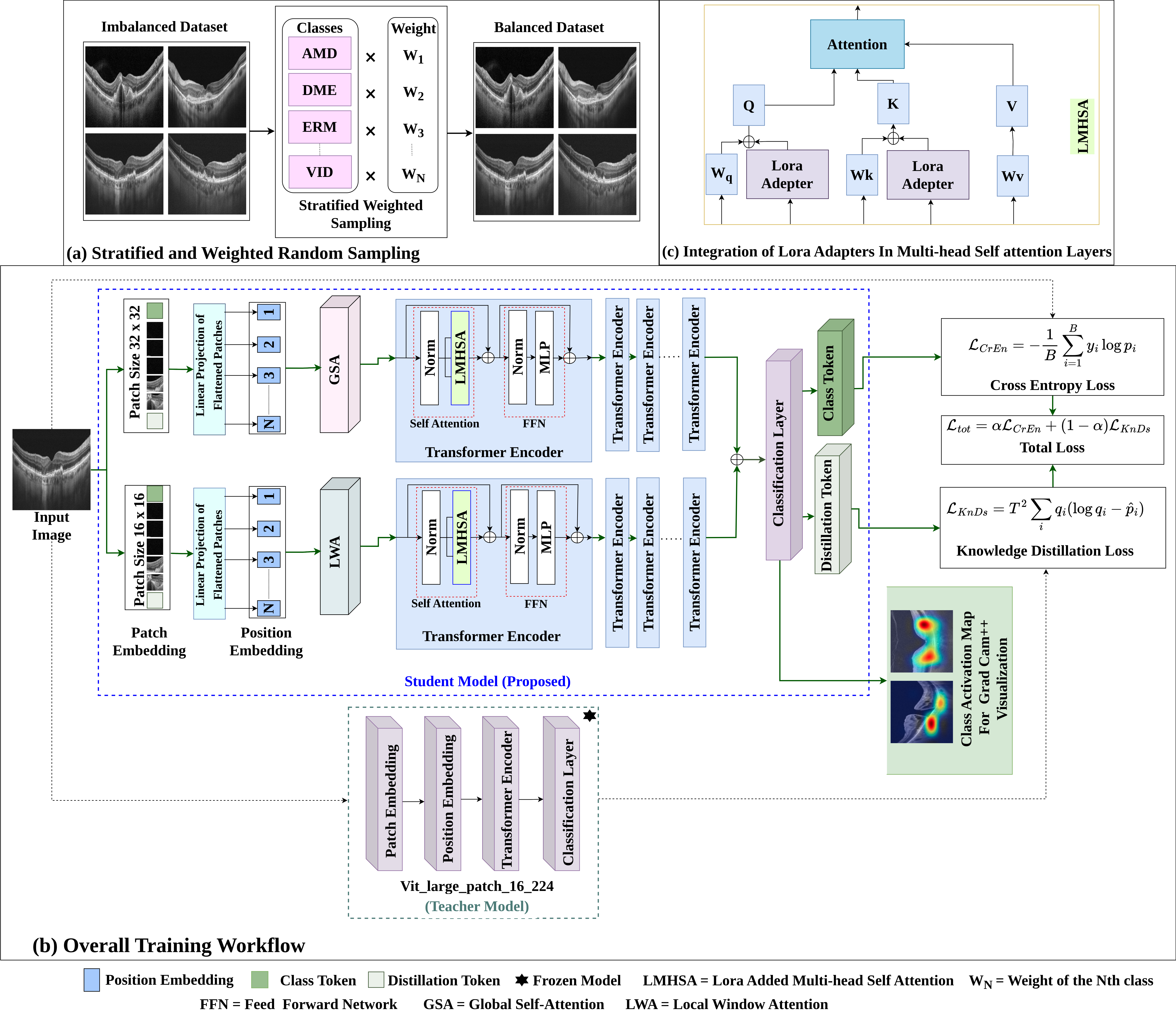}
    \caption{Overview of the Proposed Model and Training Workflow. (a) Shows stratified and weighted random sampling to balance class distribution. (b) Depicts the training pipeline with context aware multiscale patch embeddings, LoRA-integrated Transformer encoders, and a combined loss function with Grad-CAM++ visualization. (c) Highlights LoRA adapter integration within the multi-head self-attention mechanism.}
    \label{fig:Figure_2}
\end{figure*}
\begin{algorithm*}[!h]
\caption{Pseudo Code for Training the Proposed Model In Federated Setup.}  
\label{alg:algorithm_1}
\begin{algorithmic}[1]
\Require Number of clients $C$, Number of communication rounds $T$, Batch size $B$,  
         Local training dataset $\mathcal{D}_{train}^{(c)}$, Validation dataset $\mathcal{D}_{val}^{(c)}$, Test dataset $\mathcal{D}_{test}^{(c)}$, Teacher Model $\mathcal{M}_{teacher}$. 

\Ensure Trained global Proposed Model $\mathcal{M}_{student}$ with LoRA and training history.

\State \textbf{Initialize Global Models:}  
    \State Load Models $\mathcal{M}_{teacher}$ and $\mathcal{M}_{student}$  
    \State Initialize LoRA adaptation layers in $\mathcal{M}_{student}$ and set initial LoRa parameters  $\theta_{lora}^0$.

\State \textbf{Dataset:}  
    \State Prepare $\mathcal{D}_{train}^{(c)}$ using Stratified and Weighted Random Sampling among $C$ clients
\For{each round $t = 1$ to $T$}  
    \State Broadcast current student global LoRA parameters $\theta_{lora}^{(t-1)}$ to all clients
    \For{each client $c \in C$ in parallel}  
        \State Receive global LoRA parameters $\theta_{lora}^{(t-1)}$
        \For{epoch $e = 1$ to $E$(Local training epochs)}  
            \State Train local student model using local dataset $\mathcal{D}_{train}^{(c)}$
            \State Get teacher model predictions: $y_{teacher}^{(c)} \gets \mathcal{M}_{teacher}(\mathcal{D}_{train}^{(c)})$
            \State Calculate Cross Entropy Loss and Knowledge Distillation Loss  
            \State Update \textbf{only LoRA parameters} in student model  
            \State Evaluate Local Student Model on validation set:
    \[
    (\text{Loss}_{val}, \text{Accuracy}_{val}) \gets Evaluate(\mathcal{M}_{student}, \mathcal{D}_{val})
    \]
    \State Store training history $(\text{Loss}_{val}, \text{Accuracy}_{val})$
        \EndFor
    \EndFor

    \State \textbf{Aggregate client updates:}
    \State Compute weighted aggregation of LoRA parameters using FedAvg:
    \[
    \theta_{lora}^t \gets \sum_{c=1}^{C} \frac{n_c}{N} \theta_{lora, c}^t
    \]
    where $n_c$ is client dataset size, $N$ is total dataset size.

    \State Update Global Student Model with aggregated $\theta_{lora}^t$
\EndFor
\end{algorithmic}
\end{algorithm*}
\subsection{Proposed Model}
Our proposed framework adopts a FL paradigm to facilitate the training of privacy-preserving models in medical ophthalmic imaging. In this decentralized architecture, multiple client institutions independently train local instances of the proposed model on their respective datasets, ensuring that sensitive patient data remains securely within each institution's boundaries. Following local training, each client evaluates its model using a local validation set and transmits only the updated Low-Rank Adaptation (LoRA) weights to a central server. These updates are then aggregated using the Federated Averaging (FedAvg) algorithm to produce a unified global model. The global model—maintaining architectural consistency with the local models—is initialized with the aggregated LoRA parameters and subsequently evaluated on a separate test set. Its updated LoRA weights are then redistributed to the clients, initiating the next round of federated training. This iterative process continues until convergence. An overview of the federated learning workflow is presented in \hyperref[fig:Figure_1]{Figure \ref*{fig:Figure_1}}.

The architectural design and training strategy, depicted in \hyperref[fig:Figure_2]{Figure \ref*{fig:Figure_2}}, are structured to address key challenges in medical imaging, including class imbalance, computational efficiency, and model interpretability. Our analysis highlights that rare diseases within similar diagnostic categories are often underrepresented, which adversely impacts performance on minority classes. To mitigate this, we adopt a stratified sampling approach enhanced by weighted random sampling during training (\hyperref[fig:Figure_2]{Figure \ref*{fig:Figure_2}(a)}). This method increases the relative presence of underrepresented classes, thereby improving the model's sensitivity and diagnostic accuracy.

To enhance feature extraction, the model employs a context-aware multiscale patch embedding strategy using patch sizes of 16×16 and 32×32 (\hyperref[fig:Figure_2]{Figure \ref*{fig:Figure_2}(b)}). This design enables the model to effectively capture both fine-grained local details and broader contextual information. Specifically, the 16×16 patches are processed using local window attention(LWA), which focuses on preserving intricate spatial features, while the 32×32 patches are subjected to global self-attention(GSA) to encode high-level semantic representations. The outputs from these two branches are subsequently fed into the transformer encoder. By integrating information from multiple spatial resolutions, the model is better equipped to detect subtle pathological patterns, which is critical for accurate diagnosis of ophthalmic disease.

One of the major limitations in FL environments is the computational overhead associated with maintaining consistent training conditions across heterogeneous clients. To address this, we introduce a LoRA-integrated multi-head self-attention mechanism within the Transformer Encoder—denoted as LMHSA in \hyperref[fig:Figure_2]{Figure \ref*{fig:Figure_2}(b)}. LoRA significantly reduces the number of trainable parameters, making the model more efficient and well-suited for deployment in resource-constrained clinical settings. The detailed integration of LoRA adapters within the attention mechanism is presented in \hyperref[fig:Figure_2]{Figure \ref*{fig:Figure_2}(c)}.

To further enhance generalization, particularly in data-scarce environments, we employ a knowledge distillation strategy. A pre-trained ViT-Large-Patch16-224 model serves as the teacher, transferring learned representations to the lightweight student model. This student-teacher framework contributes to greater training stability and predictive consistency, as visualized in \hyperref[fig:Figure_2]{Figure \ref*{fig:Figure_2}(b)}.

Throughout the FL process, only LoRA parameters are updated and communicated between clients and the central server, while the core Transformer weights remain static. This selective update mechanism greatly reduces communication overhead. For experimentation, we simulate four decentralized clients by partitioning each dataset into non-overlapping subsets, thereby replicating the data siloing commonly found in real-world clinical deployments.

\hyperref[fig:Figure_2]{Figure \ref*{fig:Figure_2}} offers a detailed overview of the workflow, while \hyperref[alg:algorithm_1]{Algorithm \ref*{alg:algorithm_1}}  formalizes the client-server training protocol. The subsequent subsections provide an in-depth analysis of each architectural component and its contribution to overall system performance.

\subsubsection{Stratified Sampling with Weighted Random Sampling}
Handling class imbalance is a critical challenge in medical image classification, particularly when certain pathological conditions are significantly underrepresented in real-world medical imaging datasets.  To address this, we applied stratified and weighted random sampling locally on each client within the FL setup, ensuring balanced class representation during training and improving overall model robustness.

Given a dataset with $N$ number of images with $C$ classes, where each class contains $N_c$ samples, the class distribution can be defined as follows,
\begin{equation}
    P_c = \frac{N_c}{N}
\end{equation}
where $P_c$ is the proportion of class $c$ in the dataset.
Compared to standard sampling, this method alleviates the issue of minority classes being sampled less frequently and provides a proportional representation of every class.

To further refine and rectify class imbalances among the minority classes, we implemented Weighted Random Sampling, wherein each sample is allocated a selection probability determined by its corresponding class weight, which can be mathematically expressed as follows.
\begin{equation}
    w_c = \frac{N}{C \cdot N_c}
\end{equation}
Here, $w_c$  is the weight for a particular class. This weight ensures that classes with fewer instances are sampled more frequently. The probability of selecting a sample $i$ from class $c$ is then given by \citep{Nguyen2021-bb}, 
\begin{equation}
    P_i = \frac{w_{y_i}}{\sum_{j=1}^{N} w_{y_j}}
\end{equation}
here, $y_i$  is the class label of sample $i$ . This approach prevents the model from bias toward the dominant class and encourages learning from underrepresented categories. This procedure is illustrated in \hyperref[fig:Figure_2]{Figure \ref*{fig:Figure_2}(a)}.
\subsubsection{Integration of Multiscale Patch Embedding}
The input images to the transformer are first divided into smaller, non-overlapping patches, each flattened and embedded into a high-dimensional vector \citep{yin2022vit}. Patch Embedding transforms an image $x\in \mathbb{R}^{C\times H\times W}$ where $C,H,W$ are the channels, height and width from the input space $X$ into a sequence of tokens $\left\{Z_i\right\}_{i=1}^N$,  where $z_i \in \mathbb{R}^d$  is a vector in the patchification space \( Z \). To enrich the model’s ability to capture both fine-grained and coarse-grained features, we adopt a multi-scale patch embedding strategy. \hyperref[fig:Figure_2]{Figure \ref*{fig:Figure_2}(b)} illustrate two different patch embeddings for patch sizes $16\times16$ and $32\times32$. We chose $16\times16$ and $32\times32$ dimensions specifically to strike a balance between capturing fine-grained local details and understanding the broader global context. After deviding into patches they are passed through LWA and GSA. This setup enables the model to effectively identify both small pathological features and larger anatomical structures, which significantly improves its performance over the standard DeiT model that uses a single-scale embedding approach and thus lacks this level of spatial diversity.

For the two patch sizes, \( P1(16\times16) \) and \( P2(32\times32) \), used in this model, the patch embedding operation applies a convolutional layer with kernel size and stride set to \( P1 \) and \( P2 \), respectively, followed by flattening into a sequence. The operation can be expressed as\citep{Liu2025-tn}:
\begin{equation}
    {Z}_{P_1} = \text{Conv2D}(X, W_{P_1}, P_1, P_1)
\end{equation}
\begin{equation}
       Z_{P_2} = \text{Conv2D}(X, W_{P_2}, P_2, P_2)
\end{equation}
here $Z_{P_1}$ and $Z_{P_2}$ is the resulting patch embeddings for two different Patch size $P_1$ and $P_2$, $X$ is the input image, $W_{P_1}$ and $W_{P_2}$ are the two convolutional filters for patch embedding. The patch embedding sequences are subsequently extended with a classification token (CT) and a distillation token (DT), and positional embeddings are incorporated, which can be expressed as, 
\begin{equation}
    \tilde{Z}_{P_1} = [\text{CT}; \text{DT}; Z_{P_1}] + E_{P_1}
\end{equation}
\begin{equation}
    \tilde{Z}_{P_2} = [\text{CT}; \text{DT}; Z_{P_2}] + E_{P_2}
\end{equation}
here, $\tilde{Z}_{P_1}$ and $\tilde{Z}_{P_2}$ are augmented patch embeddings and $E_{P_1}$ and $E_{P_2}$ are positional embedding matrix.To further enhance the discriminative capacity of the extracted patch embeddings and emphasize clinically relevant pathological regions, we apply distinct attention mechanisms tailored to each patch scale. Specifically, the refined embedding $\tilde{Z}_{P_1}$, corresponding to finer-grained patches, is processed using a Local Self-Attention (LSA) module to capture localized feature interactions within spatially constrained windows. In parallel, the coarser-scale embedding $\tilde{Z}_{P_2}$ undergoes Global Self-Attention (GSA), enabling the model to integrate broader contextual dependencies across the entire image. This dual-attention strategy ensures that both local structural abnormalities and global contextual cues are effectively modeled, as expressed in the following equations.
\begin{equation}
    \tilde{Z}_{P_1} = \text{LWA}(\tilde{Z}{P_1})
\end{equation}
\begin{equation}
    \tilde{Z}_{P_2} = \text{GSA}(\tilde{Z}{P_2})
\end{equation}
To efficiently process extended sequences, each augmented token sequence is independently passed through a transformer encoder with multihead self attention,
\begin{equation}
H_{P_1} = \text{L}_{Trans}(\tilde{Z}{P_1})
\end{equation}
\begin{equation}
H_{P_2} = \text{L}_{Trans}(\tilde{Z}{P_2})
\end{equation}
here, $H_{P_1}$ and $H_{P_2}$ are transformer output for small and large positional embedding. To effectively combine features across scales, we extract the CLS tokens from both encoded sequences and enhance them using bidirectional linear cross-attention. That is, each CLS token attends to the token sequence of the opposite scale:
\begin{equation}
\hat{z}_{\text{CLS}1} = \text{LinearCrossAttn}(H{P_1}^{\text{CLS}}, H{P_2}^{\text{tokens}})
\end{equation}
\begin{equation}
\hat{z}_{\text{CLS}2} = \text{LinearCrossAttn}(H{P_2}^{\text{CLS}}, H{P_1}^{\text{tokens}})
\end{equation}
here, $\hat{z}_{\text{CLS}1}$ and $\hat{z}_{\text{CLS}2}$ are the updated small and large CLS token. The updated CLS tokens from both scales are concatenated and passed through a lightweight MLP for final fusion:
\begin{equation}
z_{\text{fused}} = \text{MLP}([\hat{z}_{\text{CLS}1} , \Vert , \hat{z}{\text{CLS}_2}])
\end{equation}
This fused representation $z_{\text{fused}}$ is then forwarded to the classification head.
\begin{figure*}[!h]
    \centering
    \includegraphics[width=1\linewidth]{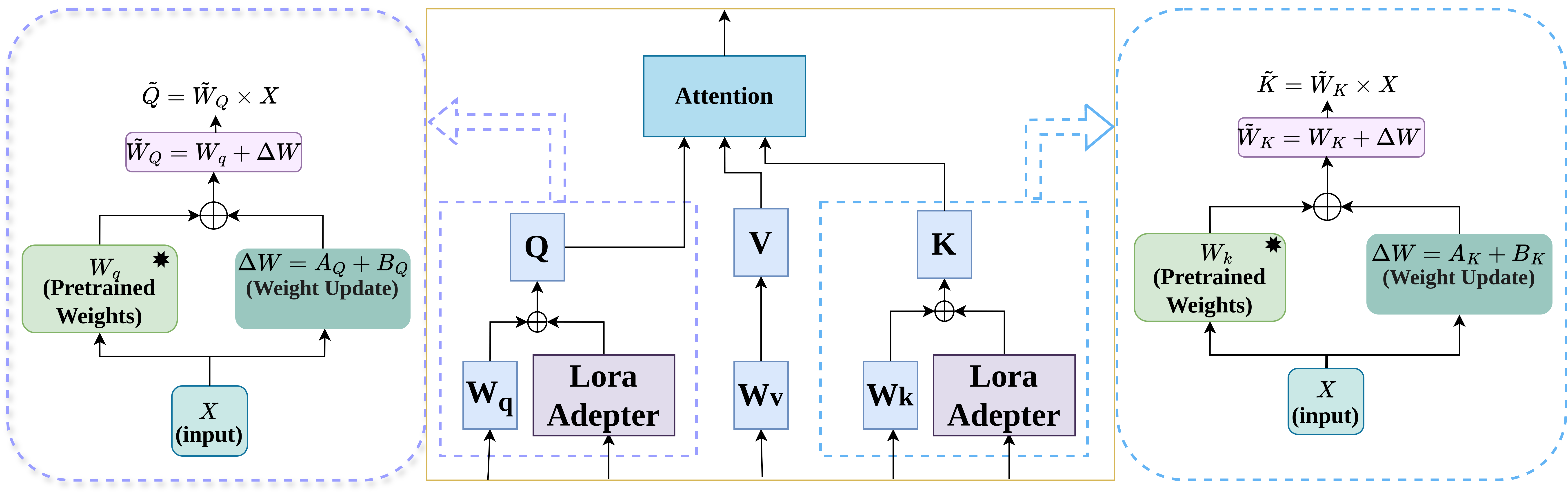}
    \caption{LoRA integration into transformer attention layers, showing low-rank updates added to frozen pretrained weights in the query and key projections.}
    \label{fig:Figure_3}
\end{figure*}
\subsubsection{LoRA Integrated Transformer Encoder}
Training Transformer models for medical image classification is computationally intensive due to their large number of parameters. Low-Rank Adaptation (LoRA) addresses this challenge by incorporating trainable low-rank matrices within the attention mechanism while keeping most of the pre-trained model unchanged \citep{Lin2024-lv}. 
We integrated LoRA into the projection layers of the query (Q) and key (K) matrices within each multi-head self-attention block, as shown in \hyperref[fig:Figure_3]{Figure \ref*{fig:Figure_3}}. LoRA introduces low-rank trainable matrices $A$ and $B$, where the rank \( r \) is much smaller than the dimension of the attention heads \( d_k \), which minimizes additional parameter overhead. This can be expressed as\citep{song2024low}, 
\begin{equation}
    \Delta W_Q = A_Q B_Q, \quad \Delta W_K = A_K B_K\quad
\end{equation}
here, $\Delta W_Q$ and $\Delta W_K$ are learnable weight matrices of $Q$ and $K$, $A_Q, A_K \in \mathbb{R}^{d \times r}$ and $ \quad B_Q, B_K \in \mathbb{R}^{r \times d_k}$. In our implementation, the rank $r$ was set to 4.
 \\
The updated query(Q) and key(K) matrices are computed as, 
\begin{align}
    \tilde{W}_Q = W_Q + \Delta W_Q = W_Q + A_Q B_Q\\
    \tilde{W}_K = W_K + \Delta W_K = W_K + A_K B_K
\end{align}
here, $ \tilde{W}_Q$ and $\tilde{W}_K$ are updated learnable weight matrices of $Q$ and $K$. The original weights $W_K$ and $W_Q$ are kept frozen as part of the pretrained transformer, allowing the model to train only the newly introduced low-rank matrices. Given an input sequence, where \( X \in \mathbb{R}^{N \times E} \), where \( N \) is the sequence length and \( E \) is the embedding dimension, the LoRA-based attention mechanism can be defined as \citep{Arnab2021-yk}, 
\begin{equation}
    \tilde{Q} = X \tilde{W}_Q, \quad \tilde{K} = X \tilde{W}_K
\end{equation}
A pictorial depiction of the LoRA integration into the attention layers has been given in \hyperref[fig:Figure_3]{Figure \ref*{fig:Figure_3}}.
\begin{equation}
    \text{LoRA-Attention}(X) = \text{Softmax} \left( \frac{\tilde{Q} \tilde{K}^T}{\sqrt{d_k}} \right) V
\end{equation}
Upon finding the $\tilde{Q}$ and $\tilde{K}$, the attention features are calculated according to Equation 19. The output of the attention layer is followed by a Feed-Forward Network (FFN), which further leads to the classification layer as,
\begin{equation}
\text{FFN}(X) = \sigma(XW_1 + b_1)W_2 + b_2
\end{equation}

where \( W_1 \in \mathbb{R}^{E \times d_k} \) and \( W_2 \in \mathbb{R}^{d_k \times E} \) are projection matrices, and \( \sigma \) is a non-linearity function. 

\subsubsection{Loss Function}
To train the proposed model, we implemented a knowledge distillation framework in which the loss is derived from the results of both the student and teacher models. In this framework, our proposed transformer model serves as the student model, while a ViT-Large-Patch16-224 acts as the pre-trained teacher model on ImageNet to leverage its high capacity and strong generalization ability, following the successful distillation strategy from DeiT. In preliminary experiments, fine-tuning the teacher on our dataset yielded marginal gains but increased the risk of overfitting, so we retained the ImageNet pre-trained. We employ cross-entropy loss to assess the discrepancy between the prediction of the model and the ground-truth labels. Given the predicted logits $zs$ from the student model, the corresponding probability distribution($p_i$) is obtained by applying the softmax function,
\begin{equation}
    p_i = \frac{e^{zs_i}}{\sum_{j} e^{zs_j}}
\end{equation}
Here, $zs$ is the logits from the student model and $p_i$ is the student predicted probability distribution.
The cross-entropy loss between the predicted probability $p_i$ and the ground truth target $y_i$ is given by,
\begin{equation}
    \mathcal{L}_{CrEn} = -\frac{1}{B} \sum_{i=1}^{B} y_i \log p_i
\end{equation}
here $\mathcal{L}_{CrEn}$ is the calculated cross entropy loss and $B$ is the batch size.

To facilitate knowledge transfer and compute the distillation loss between the teacher and student models, we employ the KL-divergence loss. The process begins by deriving the teacher model’s softened probability distribution($q_i$) using the following formulation:
\begin{equation}
    q_i = \frac{e^{zt_i / T}}{\sum_{j} e^{zt_j / T}}
\end{equation}
Here, $zt$ is the logits from the teacher model, and T is the temperature parameter for fine-tuning the knowledge distillation. In our experiments, we set T=2. The student's softened log probability distribution is calculated using, 
\begin{equation}
    \hat{p}_i = \log \left( \frac{e^{zs_i / T}}{\sum_{j} e^{zs_j / T}} \right)
\end{equation}
here, $\hat{p}_i$ is the student's softened log probability distribution. Once the teacher's softened probability distribution and the student's softened log probability distribution have been computed, the KL-divergence loss can be determined using the following formulation \citep{Kim2021-gh},
\begin{equation}
    \mathcal{L}_{KnDs} = T^2 \sum_{i} q_i (\log q_i - \hat{p}_i)
\end{equation}
here, $\mathcal{L}_{KnDs}$ is the knowledge distillation loss that measures the difference between the probability distributions generated by the teacher and student models.
Finally, the total loss can be expressed as, 
\begin{equation}
    \mathcal{L}_{tot} = \alpha \mathcal{L}_{CrEn} + (1 - \alpha) \mathcal{L}_{KnDs}
\end{equation}
Here, The total loss, represented as $\mathcal{L}_{tot}$, is the combined loss function used to train the student model., $\mathcal{L}_{CrEn}$ is the cross-entropy loss, $\mathcal{L}_{KnDs}$ is the knowledge distillation loss, and $\alpha \in [0,1]$ is the parameter which controls the trade-off between ground-truth supervision and distillation. In our case, $\alpha$ was 0.25.\\

\section{Results}
\begin{table}[ht]
\centering
\caption{Summary of Hyperparameters Used in the Proposed Model}
\begin{tabular}{|l|l|}
\hline
\textbf{Hyperparameter} & \textbf{Value / Description} \\
\hline
Batch Size & 8 \\
Learning Rate & 2e-5 \\
Image Size & 224 × 224 \\
Patch Sizes & 16 × 16 and 32 × 32 \\
LoRA Rank ($r$) & 4 \\
LoRA Alpha & 4 \\
LoRA Dropout & 0.2 \\
Optimizer & Adam \\
Optimizer Weight Decay & 1e-5 \\
Knowledge Distillation Alpha ($\alpha$) & 0.25 \\
Distillation Temperature ($T$) & 2 \\
Federated Learning Clients & 4 (simulated) \\
\hline
\end{tabular}
\label{table:Table_2}
\end{table}

\begin{table*}[ht]
\centering
\caption{\bf Evaluation Results for OCTDL Dataset on Test Data.}
\begin{tabular}{l|l|l|l|l|l|l}
\hline
\bf Model & \bf AUC$(\uparrow)$ & \bf F1$(\uparrow)$ & \bf Loss$(\downarrow)$ & \bf Precision$(\uparrow)$ & \bf Recall$(\uparrow)$ & \bf Top-5 Acc$(\uparrow)$ \\ 
\hline
RegNet & 85.36\% & 59.73\% & 1.2738 & 64.97\% & 46.85\% & 94.55\% \\
CvT & 86.30\% & 61.79\% & 1.3205 & 68.62\% & 41.30\% & 94.87\% \\
SegFormer & 89.93\% & 65.53\% & 1.0694 & 71.73\% & 56.69\% & 96.97\% \\
ViTMAE & 90.52\% & 64.34\% & 1.1168 & 79.90\% & 46.35\% & 80.34\% \\
ConvNeXt V2 & 91.38\% & 66.06\% & 1.0112 & 80.08\% & 56.45\% & 81.85\% \\
DINOv2 & 97.32\% & 81.98\% & 0.5397 & 86.10\% & 79.33\% & 91.42\% \\
ViT & 98.11\% & 84.98\% & 0.4492 & 91.50\% & 82.03\% & 99.74\% \\
Swin Transformer & 98.36\% & 89.27\% & 0.4182 & 94.38\% & 85.22\% & 99.65\% \\
EfficientFormer & 98.61\% & 85.74\% & 0.381 & 92.73\% & 84.21\% & 100.00\% \\
DeiT(Without Multiscale Path Embedding) & 99.15\% & 91.21\% & 0.3438 & 95.15\% & 86.08\% & 99.61\% \\
DeiT-x(Without LoRA) & 99.22\% & 92.38\% & 0.2659 & 95.39\% & 89.61\% & 99.88\% \\
Proposed Model & \textbf{99.24\%} & \textbf{99.18\%} & \textbf{0.0459} & \textbf{99.12\%} & \textbf{99.03\%} & \textbf{100.00\%} \\
\hline
\end{tabular}
\label{table:Table_3}
\end{table*}
\begin{table*}
\centering
\caption{\bf Evaluation Results for The Eye Disease Image Dataset on Test Data.}
\begin{tabular}{l|l|l|l|l|l|l}
\hline
\bf Model & \bf AUC$(\uparrow)$ & \bf F1$(\uparrow)$ & \bf Loss$(\downarrow)$ & \bf Precision$(\uparrow)$ & \bf Recall$(\uparrow)$ & \bf Top-5 Acc$(\uparrow)$  \\ 
\hline
CvT & 77.95\% & 41.05\% & 1.584 & 49.95\% & 11.38\% & 95.41\% \\
RegNet & 84.25\% & 51.20\% & 1.399 & 64.27\% & 16.40\% & 98.08\% \\
ConvNeXt V2 & 84.75\% & 51.52\% & 1.3506 & 57.19\% & 28.95\% & 75.63\% \\
SegFormer & 86.73\% & 52.54\% & 1.2413 & 60.63\% & 28.79\% & 99.20\% \\
ViTMAE & 88.50\% & 52.99\% & 1.2562 & 83.48\% & 23.78\% & 72.58\% \\
DINOv2 & 96.93\% & 79.46\% & 0.6032 & 83.12\% & 73.98\% & 95.38\% \\
Efficient Former & 96.96\% & 84.37\% & 0.5481 & 88.98\% & 81.33\% & 98.68\% \\
DeiT(Without Multiscale Path Embedding) & 97.20\% & 80.96\% & 0.5159 & 85.71\% & 79.52\% & 100.00\% \\
Swin Transformer & 97.92\% & 84.87\% & 0.4959 & 85.98\% & 83.35\% & 99.09\% \\
ViT & 97.97\% & 80.12\% & 0.4976 & 86.75\% & 77.21\% & 99.64\% \\

DeiT-x(Without LoRA) & 98.53\% & 87.01\% & 0.411 & 88.07\% & 81.58\% & 100.00\% \\
Proposed Model & \textbf{98.54\%} & \textbf{97.99\%} & \textbf{0.071} & \textbf{98.19\%} & \textbf{93.22\%} & \textbf{100.00\%} \\
\hline
\end{tabular}
\label{table:Table_4}
\end{table*}
The proposed model was evaluated against various architectures—ViT, DeiT, DeiT-x, RegNet, CvT, SegFormer, ViTMAE, ConvNeXt V2, Swin Transformer, DINOv2, and EfficientFormer—on the OCTDL dataset, with additional validation on the Eye Disease Dataset. All models were built in PyTorch, trained on ImageNet, and fine-tuned using the same augmentation strategies within a shared FL setup, and the hyperparameters used for this study are provided in  \hyperref[table:Table_2]{Table \ref*{table:Table_2}}. A baseline DeiT model without multiscale patch embedding or LoRA was used, followed by a DeiT-x variant with multiscale patching, and finally the proposed model with both enhancements. To reduce overfitting from oversampling, early stopping based on validation loss was applied.  \hyperref[table:Table_3]{Table \ref*{table:Table_3}} presents the model's performance on the OCTDL test data, while  \hyperref[table:Table_4]{Table \ref*{table:Table_4}} details its assessment on the Eye Disease Data set test data. The validation performance of the proposed model was assessed on the OCTDL and Eye Disease datasets, as illustrated in  \hyperref[table:Table_5]{Tables \ref*{table:Table_5}} and  \hyperref[table:Table_6]{\ref*{table:Table_6}}, respectively.

\begin{table*}
\centering
\caption{\bf Evaluation Results for  OCTDL Dataset on Validation Data. }
\begin{tabular}{l|l|l|l|l|l|l}
\hline
\bf Model & \bf AUC$(\uparrow)$ & \bf F1$(\uparrow)$ & \bf Loss$(\downarrow)$ & \bf Precision$(\uparrow)$ & \bf Recall$(\uparrow)$ & \bf Top-5 Acc$(\uparrow)$  \\ 
\hline
CvT & 85.74\% & 59.71\% & 1.2959 & 68.80\% & 41.75\% & 93.69\% \\
RegNet & 86.98\% & 61.17\% & 1.2073 & 65.24\% & 51.94\% & 97.09\% \\
ViTMAE & 87.94\% & 59.71\% & 1.2256 & 73.68\% & 40.78\% & 78.64\% \\
ConvNeXt V2 & 87.23\% & 66.02\% & 1.1947 & 75.00\% & 50.97\% & 78.16\% \\
SegFormer & 89.30\% & 63.59\% & 1.1179 & 68.71\% & 54.37\% & 95.63\% \\
DINOv2 & 96.68\% & 84.87\% & 0.628 & 86.57\% & 80.50\% & 93.00\% \\
EfficientFormer & 96.99\% & 84.47\% & 0.5638 & 87.17\% & 79.13\% & 98.54\% \\
ViT & 97.98\% & 86.41\% & 0.4602 & 91.30\% & 81.55\% & 100.00\% \\
DeiT(Without Multiscale Path Embedding) & 97.94\% & 84.47\% & 0.488 & 88.65\% & 79.61\% & 99.03\% \\
Swin Transformer & 97.77\% & 83.50\% & 0.492 & 88.17\% & 79.61\% & 98.06\% \\
DeiT-x(Without LoRA) & 98.28\% & 84.47\% & 0.4198 & 89.36\% & 81.55\% & 100.00\% \\
Proposed Model & \textbf{98.39\%} & \textbf{98.54\%} & \textbf{0.0432} & \textbf{98.54\%} & \textbf{98.54\%} & \textbf{100.00\%} \\
\hline
\end{tabular}
\label{table:Table_5}
\end{table*}
\begin{table*}
\centering
\caption{\bf Evaluation Results for The Eye Disease Image Dataset on Validation Data.}
\begin{tabular}{l|l|l|l|l|l|l}
\hline
\bf Model & \bf AUC$(\uparrow)$ & \bf F1$\uparrow$ & \bf Loss$(\downarrow)$ & \bf Precision$(\uparrow)$ & \bf Recall$(\uparrow)$ & \bf Top-5 Acc$(\uparrow)$ \\ 
\hline
CvT & 80.78\% & 41.55\% & 1.5232 & 31.03\% & 6.34\% & 98.59\% \\
RegNet & 85.67\% & 50.70\% & 1.3077 & 65.79\% & 17.61\% & 95.77\% \\
ConvNeXt V2 & 86.67\% & 52.11\% & 1.2515 & 55.41\% & 28.87\% & 75.30\% \\
SegFormer & 87.77\% & 52.82\% & 1.2016 & 67.61\% & 33.80\% & 98.59\% \\
ViTMAE & 87.93\% & 50.00\% & 1.2781 & 79.31\% & 16.20\% & 73.94\% \\
DINOv2 & 97.81\% & 85.92\% & 0.4473 & 88.06\% & 83.10\% & 96.48\% \\
Efficient Former & 98.33\% & 81.69\% & 0.449 & 85.04\% & 76.00\% & 100.00\% \\
Swin Transformer & 98.55\% & 88.03\% & 0.4245 & 90.00\% & 82.39\% & 100.00\% \\
DeiT-x(Without LoRA) & 98.64\% & 88.73\% & 0.382 & 88.24\% & 84.51\% & 100.00\% \\
DeiT(Without Multiscale Path Embedding) & 98.73\% & 87.32\% & 0.3729 & 88.49\% & 86.62\% & 100.00\% \\
ViT & 99.14\% & 88.73\% & 0.3568 & 92.31\% & 84.51\% & 100.00\% \\
Proposed Model & \textbf{99.45\%} & \textbf{98.28\%} & \textbf{0.0446} & \textbf{98.27\%} & \textbf{94.25\%} & \textbf{100.00\%} \\
\hline
\end{tabular}
\label{table:Table_6}
\end{table*}

Across both datasets, the proposed model consistently outperformed all other models in classification accuracy, robustness, and generalization. On the OCTDL test set, it achieved the highest AUC (99.24\%), F1 score (99.18\%), and precision (99.12\%) while maintaining the lowest cross-entropy loss (0.0459). The Top-5 accuracy of 100\% further emphasized its classification reliability. While DeiT-x achieved a similar AUC, its F1 score and loss were notably less favorable. Transformer-based models generally outperformed convolution-based alternatives; ConvNeXt V2 and EfficientFormer showed reduced precision and F1 scores, and RegNet and CvT underperformed consistently across all key metrics, suggesting that convolutional backbones may struggle with complex global feature interactions present in ophthalmic imagery. The results are provided in  \hyperref[table:Table_3]{Table \ref*{table:Table_3}}.

Validation results mirrored the test performance, underscoring the model’s generalization capability. On the OCTDL validation set, our model again achieved the top AUC (98.39\%) with high F1 and precision scores and the lowest loss (0.0432) illustrated in  \hyperref[table:Table_5]{Table \ref*{table:Table_5}}. Comparable superiority was observed on the Eye Disease dataset, where the model reached a validation AUC of 99.45\% and maintained a Top-5 accuracy of 100\%. This consistent validation performance, particularly in data-scarce medical scenarios, highlights the robustness of the model’s training strategy and architecture design. The results are provided in  \hyperref[table:Table_6]{Table \ref*{table:Table_6}}.

\begin{figure*}[!ht]
    \centering
    \includegraphics[width=1\linewidth]{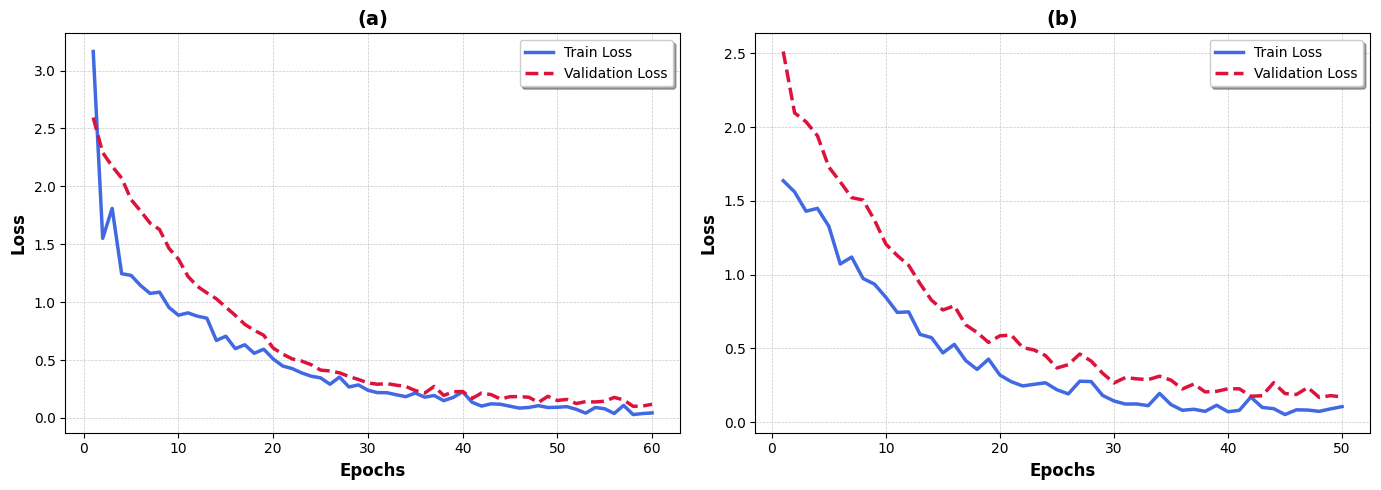}
    \caption{Training and validation performance over epochs for two datasets: (a) OCTDL and (b) Eye Disease.}
    \label{fig:Figure_4}
\end{figure*}
\begin{figure*}
    \centering
    \includegraphics[width=0.8\linewidth]{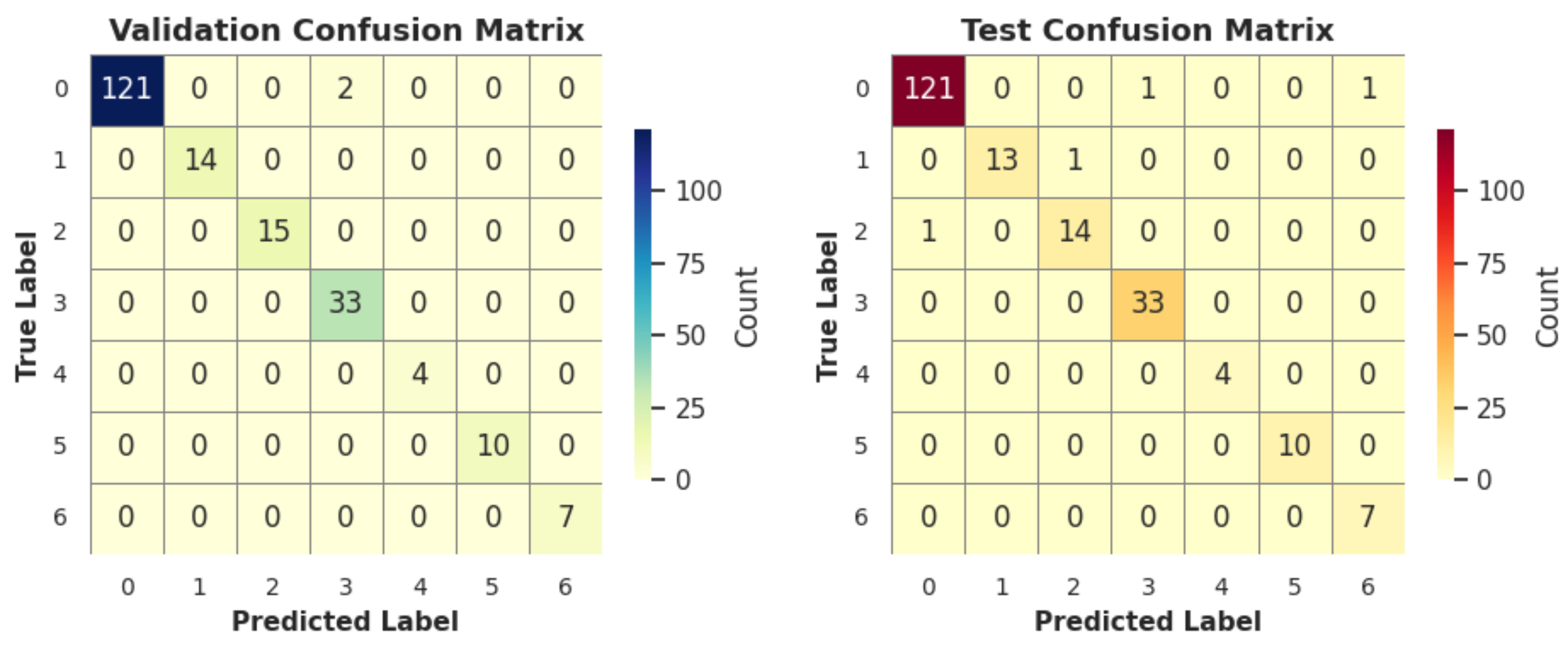}
    \caption{Confusion Metrics for OCTDL Dataset}
    \label{fig:Figure_5}
\end{figure*}
\begin{figure*}[!h]
    \centering
    \includegraphics[width=0.8\linewidth]{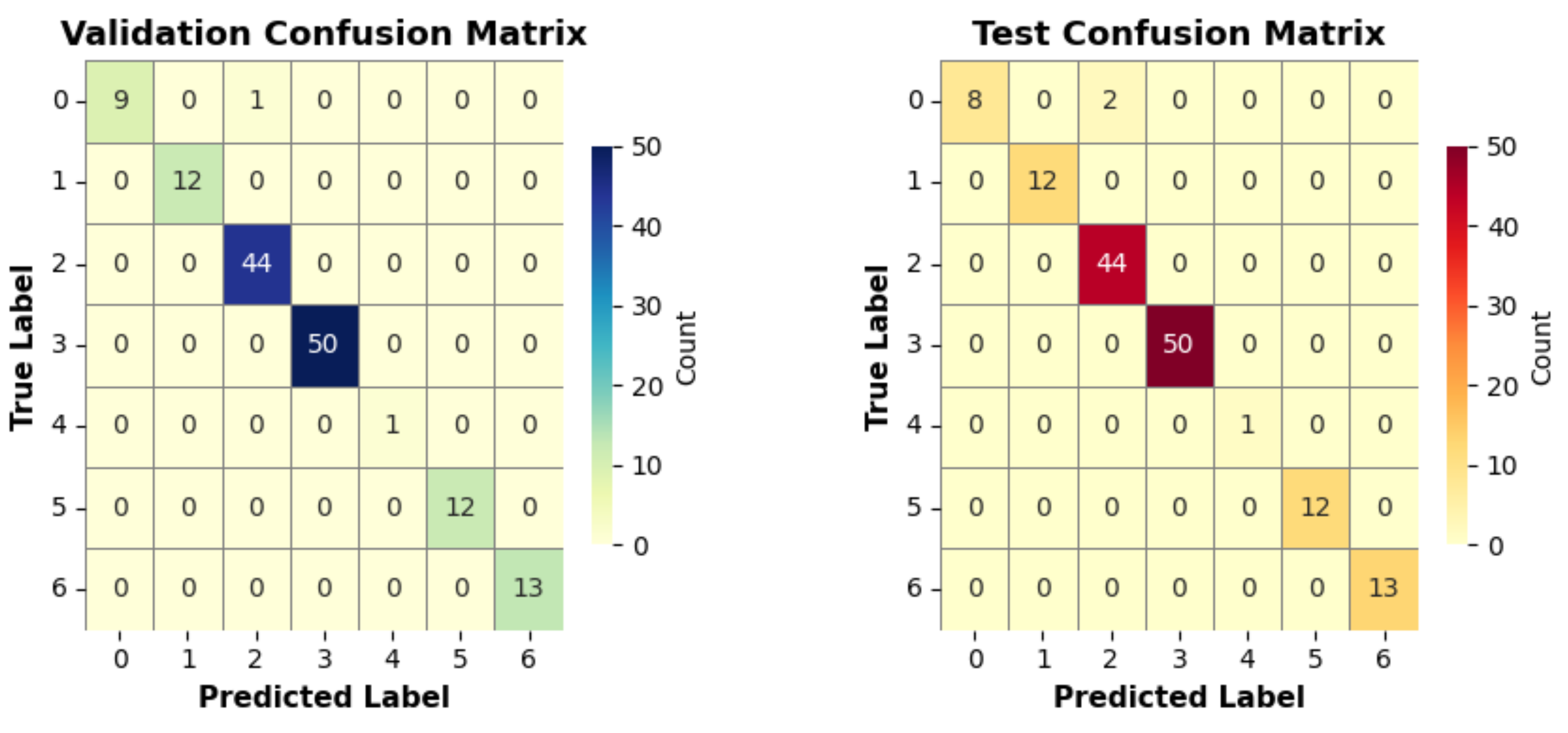}
    \caption{Confusion Metrics for Eye Disease Dataset}
    \label{fig:Figure_6}
\end{figure*}
\begin{figure*}[!h]
    \centering
    \includegraphics[width=0.5\linewidth]{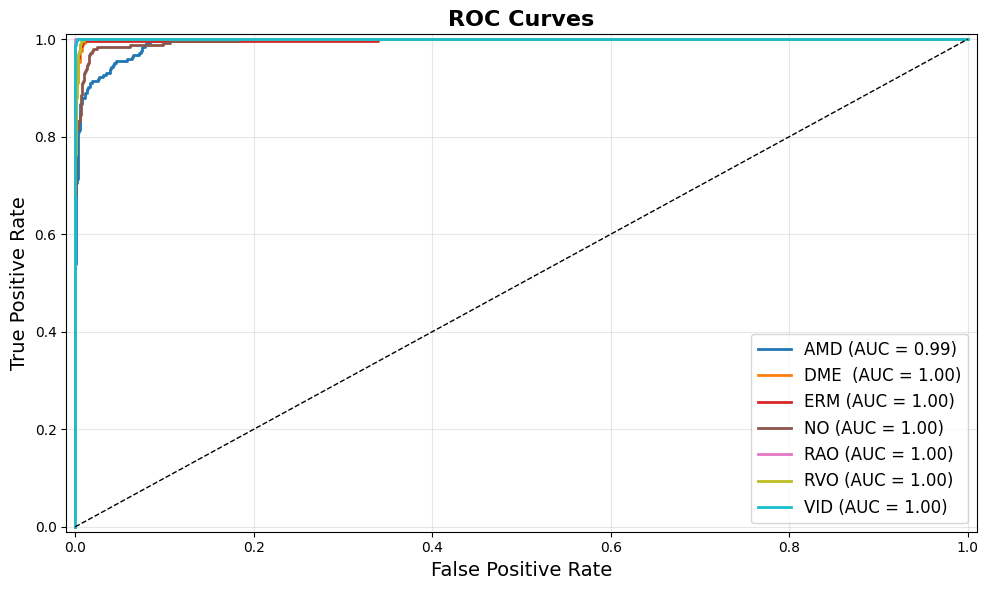}
    \caption{ROC Curve for our Proposed Model on OCTDL Test Data}
    \label{fig:Figure_7}
\end{figure*}
\begin{figure*}[!h]
    \centering
    \includegraphics[width=0.5\linewidth]{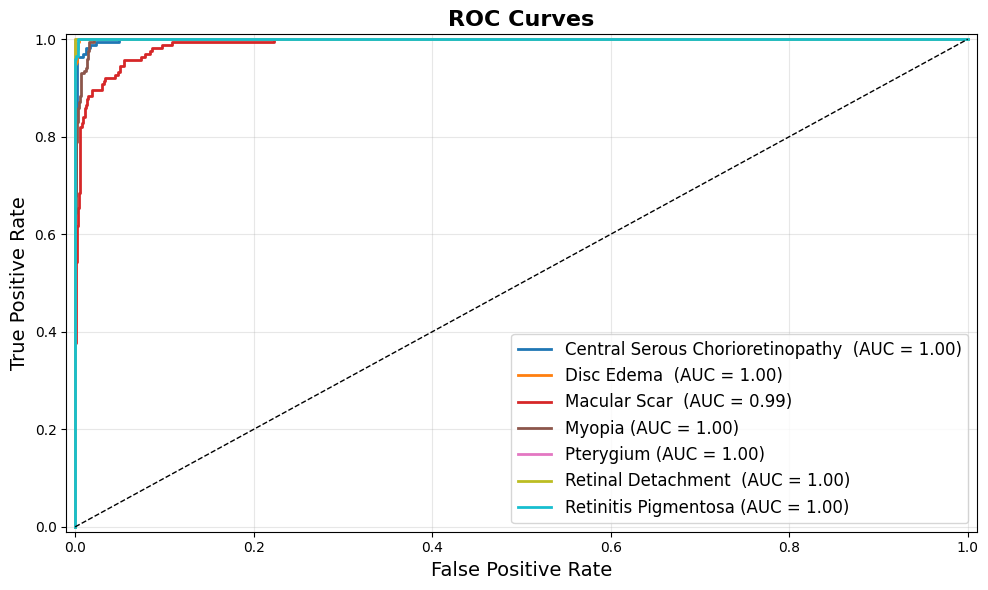}
    \caption{ROC Curve for our Proposed Model on Eye Disease Dataset Test Data}
    \label{fig:Figure_8}
\end{figure*}
\begin{figure*}[!ht]
    \centering
    \includegraphics[width=0.7\linewidth]{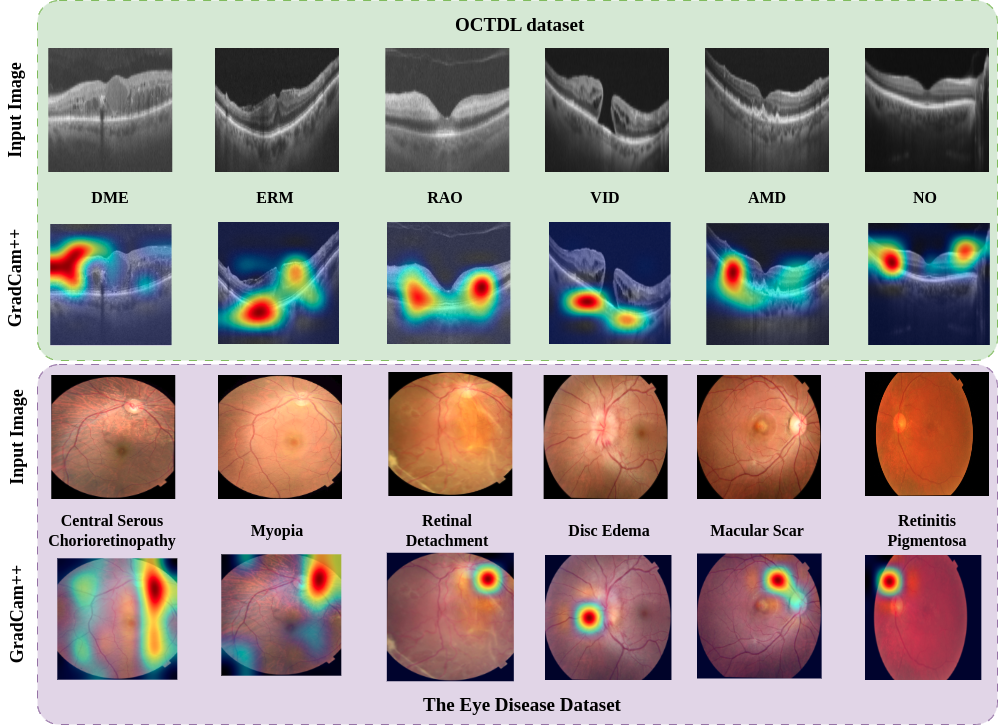}
    \caption{Grad Cam++ Visualization of Various Classes}
    \label{fig:Figure_9}
\end{figure*}
\begin{table*}
\centering
\caption{\bf Comparison of the proposed model against state-of-the-art works using the OCTDL and EDD datasets.}
\begin{tabular}{l|p{3cm}|p{2cm}|l|l|l|l|l}
\hline
\bf Ref& \bf Venue(Year) &\bf Model& \bf Dataset & \bf AUC$(\uparrow)$ & \bf F1$\uparrow$ & \bf Precision$(\uparrow)$ & \bf Recall$(\uparrow)$  \\ 
\hline
\cite{shamsan2023automatic}& Diagnostics(2023)&MobileNet and DenseNet121 & EDD &  99.23\% &-&98.45\%& 98.75\%\\
\cite{muthukannan2022optimized}&Computers in Biology and Medicine(2022)&CNN-MDD& OCTDL & -&93.3\%&98.30\%&95.21\%\\
\cite{csener2023classification}&ELECO(2023)&EfficientNetB0, VGG-16, and VGG-19& EDD & 97.94\% & - & 93.92\% & 93.90\%\\
\cite{ouda2022multiple}&Electronics(2022)&ML-CNN& EDD & 96.7\% & - & 91.5\% &
80\%\\
\cite{sharma2024efficient}&Engineering, Technology \& Applied Science Research(2024)&EfficientNet-B5 & EDD & -& 96.02\% & 96.08 \% & 96.04 \%\\
\cite{he2023interpretable}&Scientific Reports(2023)&Swin-Poly Transformer& OCTDL & - &97.10\% & 97.13\% & 97.13\%\\
\cite{jaimes2025detection}&Discover Applied Sciences(2025)&VGG16 & OCTDL & - & 95.20\% & 95.29\% & 95.19\%\\
\cite{agarwal2025hdl}& Scientific Reports(2025)& HDL-ACO& OCTDL & - & 92.4\% & 92.8\% & 	
92.1\% \\
\textbf{Ours}&-& Context-aware multi-patch embedded DeiT & OCTDL+EDD  &\textbf{99.24\%} & \textbf{99.18\%} & \textbf{99.12\%} & \textbf{99.03\%} \\
\hline
\end{tabular}
\label{table:Table_7}
\end{table*}
The convergence behavior and optimization stability of the model were further analyzed through training and validation loss curves, as shown in \hyperref[fig:Figure_4]{Figure \ref*{fig:Figure_4}}. The loss profiles over 60 epochs for OCTDL and 50 epochs for the Eye Disease dataset demonstrate smooth convergence. An early stopping strategy with a patience of 10 epochs was employed to avoid overtraining, striking a balance between performance and generalization. This mechanism effectively prevented both premature termination and excessive fitting, ensuring reliable optimization across datasets.

Further evaluation using confusion matrices (\hyperref[fig:Figure_5]{Figure \ref*{fig:Figure_5}}and \hyperref[fig:Figure_6]{\ref*{fig:Figure_6}}) provided a granular view of the model’s predictive behavior across different disease categories. The matrices for both test and validation splits showed high true positive rates and low inter-class confusion, particularly among visually similar conditions. This result underscores the discriminative power of the model, enabled by its multiscale patch embedding and LoRA-based architecture, which effectively captures both local and global retinal features.

To assess the performance of our proposed model in classifying various ophthalmic diseases, we generated class-wise ROC curves using the held-out test sets from both the OCTDL dataset (\hyperref[fig:Figure_7]{Figure \ref*{fig:Figure_7}}) and the Eye Disease dataset (\hyperref[fig:Figure_8]{Figure \ref*{fig:Figure_8}}). These ROC curves depict the relationship between the true positive rate (sensitivity) and the false positive rate (1-specificity) for each individual class. Remarkably, the model achieved a near AUC score of 1.00 almost across all categories, reflecting its exceptional ability to distinguish between disease classes. This outcome indicates that the model is highly effective in correctly identifying both the presence and absence of each condition, underscoring its robustness and diagnostic reliability.

The Grad-CAM++ visualizations for various ophthalmic diseases can be seen in \hyperref[fig:Figure_9]{Figure \ref*{fig:Figure_9}}. The first and third rows display the input images. In contrast, the second and fourth rows illustrate the corresponding Grad-CAM++ activation maps, highlighting the discriminative regions the proposed model utilizes for classification. Warmer(red and yellow) colors indicate regions of greater significance in the model's decision-making process, whereas cooler (blue) colors correspond to less influential areas. These visualizations demonstrate the ability of the model to focus on pathology-relevant regions, thus enhancing interpretability and reinforcing its clinical reliability.

The proposed model's superior performance across diverse evaluation metrics, its consistent generalization on unseen data, and its ability to provide interpretable predictions collectively position it as a strong candidate for reliable and scalable ophthalmic disease classification. Unlike prior approaches, which either emphasize accuracy at the cost of model size and complexity or focus on privacy without interpretability, our model delivers across all fronts—performance, efficiency, privacy, and explainability. These findings support its potential for integration into clinical diagnostic workflows and decentralized medical imaging systems.

The proposed model consistently outperforms architectures, including ViT, DeiT, Swin Transformer, and ConvNeXt V2, across multiple evaluation metrics on both the OCTDL and Eye Disease Image datasets. The model demonstrates exceptional generalization capabilities by achieving the highest AUC, F1 scores, and precision while maintaining the lowest loss values. Furthermore, its 100\% Top-5 Accuracy reinforces its robustness in classification tasks. The analysis of training and validation loss curves confirms the effectiveness of the model’s optimization process. At the same time, the confusion matrices offer valuable insights into their predictive accuracy across various disease categories. Additionally, Grad-CAM++ visualizations enhance the model’s interpretability by localizing pathology-relevant regions, thereby increasing its clinical reliability. Collectively, these findings establish the proposed model’s superiority in ophthalmic disease classification, underscoring its potential for reliable and automated medical image analysis.

\section{Discussions}
The experimental results robustly validate the effectiveness of our proposed model in addressing the complex challenges of ophthalmic disease classification. By leveraging a context-aware multiscale patch embedding strategy, the model adeptly captures both fine-grained local abnormalities and global anatomical structures, which are critical for diagnosing retinal diseases. This multi-scale representation, coupled with LoRA-enhanced transformer encoders, facilitates superior feature extraction without incurring significant computational overhead. Furthermore, the use of federated learning ensures data privacy across decentralized institutions, while knowledge distillation improves generalization from limited datasets. The consistent dominance of our model across all evaluation metrics—especially in terms of F1 score, AUC, and loss minimization—reflects a well-balanced optimization between model complexity and diagnostic reliability. Additionally, the Grad-CAM++ visualizations offer strong interpretability, reinforcing the model's clinical transparency and trustworthiness.

Compared with other models evaluated in our results section, the proposed framework demonstrates marked superiority in a range of critical performance dimensions. Unlike traditional transformer variants and CNN-based models, our approach achieves a balanced trade-off between precision, recall, and loss minimization, effectively mitigating the limitations posed by class imbalance and insufficient spatial feature representation. Architectures such as ViT, Swin Transformer, and ConvNeXt V2, while capable in isolated contexts, fail to maintain consistent performance across diverse datasets and metrics. The performance of our model is not only more stable but also more clinically reliable, particularly when recognizing minority class conditions, a challenge for most competing models. The improved optimization trajectory and minimized validation loss curves further underscore the robustness of our training methodology, reflecting effective convergence and resilience to overfitting.

Furthermore, a comparison with state-of-the-art methods as summarized in  \hyperref[table:Table_7]{Table \ref*{table:Table_7}} highlights the competitive edge of our model in real-world applicability. Although previous work has focused individually on accuracy, model interpretability, or data privacy, our model synthesizes all of these aspects into a unified framework. It consistently surpasses previous studies in AUC, F1 score, precision, and recall, demonstrating not only incremental but also holistic advancement. The ability to maintain high diagnostic performance across both the OCTDL and Eye Disease datasets, coupled with its federated design and clinical interpretability, affirms its potential as a state-of-the-art solution for scalable, secure, and explainable AI in ophthalmology.

\section{Conclusion}
This work presents a powerful, decentralized, and privacy-conscious transformer-based framework for classifying ophthalmic diseases. It incorporates context-aware multiscale patch embeddings, LoRA-enhanced attention mechanisms, federated learning, and knowledge distillation. The model successfully captures both fine-grained local details and broader global retinal patterns, delivers high performance across critical evaluation metrics, and safeguards patient data through decentralized training. Additionally, Grad-CAM++ visualizations improve interpretability, reinforcing clinical confidence in the model’s decisions. Evaluated on two benchmark datasets, the proposed model consistently outperforms existing architectures and state-of-the-art methods, demonstrating superior generalization, optimization stability, and diagnostic reliability. Unlike prior approaches that address isolated challenges, our framework offers a unified, scalable solution for secure and explainable medical AI. Future research can expand this framework by incorporating more diverse, real-world datasets and multimodal data sources, extending its effectiveness in clinical settings.


\section*{Declaration of competing interest}
We do not have any conflict of interest.

\printcredits

\bibliographystyle{elsarticle-num}
\bibliography{References}



\end{document}